\documentclass[11pt,letterpaper]{article}
\usepackage{authblk}
\usepackage{emnlp2017}
\usepackage{times}
\usepackage{latexsym}

\emnlpfinalcopy

\def\emnlppaperid{286}

\title{Using Argument-based Features to Predict and Analyse Review Helpfulness}

\author[1,2]{Haijing Liu}
\author[1]{Yang Gao}
\author[1]{Pin Lv}
\author[1,2]{Mengxue Li}
\author[3]{Shiqiang Geng}
\author[1,2]{Minglan Li}
\author[4]{Hao Wang}

\affil[1]{Institute of Software, Chinese Academy of Sciences}
\affil[2]{University of Chinese Academy of Sciences}
\affil[3]{School of Automation, Beijing Information Science and Technology University}
\affil[4]{Qihoo 360 Search Lab}

\affil[ ]{\textit {\{haijing,mengxue,minglan\}2015@iscas.ac.cn}}
\affil[ ]{\textit {\{gaoyang,lvpin\}@iscas.ac.cn, \{gsq\_study,cashenry\}@126.com}}

\date{}

\begin{document}

\maketitle

\begin{abstract}
We study the helpful product reviews identification problem in
this paper. We observe that the evidence-conclusion
discourse relations, also known as \emph{arguments},
often appear in product reviews, and
we hypothesise that some \emph{argument}-based
features, e.g. the percentage of argumentative
sentences, the evidences-conclusions ratios,
are good indicators of helpful reviews. To validate this hypothesis, we manually annotate
arguments in 110 hotel reviews,
and investigate the effectiveness of several
combinations of argument-based features.
Experiments suggest that, when being used together
with the argument-based features, the
state-of-the-art baseline features
can enjoy a performance boost
(in terms of F1) of 11.01\% in average.
\end{abstract}

\section{Introduction}

\label{sec:intro}
Product reviews have significant influences on
potential customers' opinions and their purchase
decisions \cite{chatterjee2001online,
chen2004impact, dellarocas2004exploring}.
Instead of reading a long list of reviews,
customers usually are only willing to
view a handful of \emph{helpful reviews} to make
their purchase decisions.
In other words, helpful reviews have even greater
influences on the potential customers' decision-making
processes and thus on the sales;
as a result, the automatic identification
of helpful reviews has received
considerable research attentions in recent years
\cite{kim2006automatically,liu2008modeling,mudambi2010makes,xiong2014empirical,martin2014prediction,
yang2015semantic,yang2016aspect}.

Existing works on helpful reviews identification mostly
focus on designing efficient features.
Widely used features include \emph{external features},
(e.g. date \cite{liu2008modeling}, product rating \cite{kim2006automatically} and product type \cite{mudambi2010makes})
and \emph{intrinsic features} (e.g. semantic dictionaries
\cite{yang2015semantic} and emotional dictionaries \cite{martin2014prediction}).
Compared to external features, intrinsic features can provide some insights and
explanations for the prediction results, and support
better cross-domain generalisation. In this work, we investigate a new form of intrinsic
features: the \emph{argument} features.

An argument is a basic unit people use to persuade their
audiences to accept a particular state of affairs \cite{eckle2015role}.
An argument usually consists of a \emph{claim} (also known
as \emph{conclusion}) and some
\emph{premises} (also known as \emph{evidences})
offered in support of the claim.
For example, consider the following review excerpt: ``\emph{The staff were amazing, they went out of their way to help us}'';
the texts before the comma constitute a claim, and the texts
after the comma give a premise supporting the claim.
Argumentation mining \cite{moens2013argumentation,lippi2016argumentation} receives growing research interests in various domains \cite{palau2009argumentation,contractor2012using,park2014identifying,madnani2012identifying,kirschner2015linking,wachsmuth2014modeling,wachsmuth2015sentiment}.
Recent advances in automatic arguments identification \cite{stab2014identifying},
has stimulated the usage of argument features
in multiple domains, e.g. essay scoring \cite{wachsmuth2016using}
and online forum comments ranking \cite{wei122016post}.

The motivation of this work is a hypothesis
that, the helpfulness of a review is closely related
to some argument-related features, e.g. the percentage of argumentative sentences, the average number of premises in each argument, etc.
To validate our hypothesis,
we manually annotate arguments in 110
hotel reviews so as to use these ``ground
truth''
arguments to testify the effectiveness
of argument-based features for detecting helpful
hotel reviews.
Empirical results suggest that,
for four baseline feature sets we test,
their performances can be improved, in average,
by 11.01\% in terms of F1-score and 10.40\% in terms of
AUC when they are used together with some argument-based features.
Furthermore, we use the effective argument-based features
to give some insights into which product reviews
are more helpful.

\section{Corpus}
\label{sec:corpus}
We use the Tripadvisor hotel reviews corpus
built by \cite{o2010classification} to
test the performance of our helpful reviews classifier.
Each entry in this corpus includes the
review texts, the number of people that have viewed this review
(denoted by Y) and the number of people that think this review
is helpful (denoted by X).

We randomly sample 110 hotel
reviews from this corpus to annotate the
``ground truth'' argument structures \footnote{The annotated corpus can be obtained by contacting the first author}. In line with \cite{wachsmuth2015sentiment},
we view each sub-sentence in the review as a \emph{clause} and ask three
annotators independently to annotate each clause as one of the following seven
argument components:

{\bfseries \emph{Major Claim:}}
a summary of the main opinion of a review. For instance, \emph{``I have enjoyed the stay in the hotel'', ``I am sad to say that i am very disappointed with this hotel''};

{\bfseries \emph{Claim:}}
a subjective opinion on a certain aspect of a hotel. For example, \emph{``The
staff was amazing'', ``The room is spacious''};

{\bfseries \emph{Premise:}}
an objective reason/evidence supporting a claim. For instance, \emph{``The staff went out
of their way to help us''}, it supports the first example claim above; \emph{``We had a sitting room as well as a balcony''}, it supports the second example claim above;

{\bfseries \emph{Premise Supporting an Implicit Claim (PSIC):}}
an objective reason/evidence that supporting an \emph{implicit claim}, which
does appear in review. For instance, \emph{``just five minutes' walk to the
down town''} supports some implicit claims like
\emph{``the location of the hotel is good''}, although this implicit
claims has never appeared in the review;

{\bfseries \emph{Background:}}
an objective description that does not give direct opinions but provides some
background information. For example, \emph{``We checked into this hotel at midnight'', ``I stayed five nights at this hotel because i was attending a conference at the hotel''};

{\bfseries \emph{Recommendation:}}
a positive or negative recommendation for the hotel. For instance, \emph{``I would definitely come to this hotel again the next time I visit London'', ``Do not come to this hotel if you look for some clean places
to live''};

{\bfseries \emph{Non-argumentative:}} for all the other clauses.

We use the Fleiss' kappa metric \cite{fleiss1971measuring}
to evaluate the quality of the obtained annotations,
and the results are presented in Table \ref{table:corpus}.
We can see that the lowest Kappa scores
(for Premise) is still above 0.6,
suggesting that the quality of the annotations are \emph{substantial}
\cite{landis1977measurement}; in other words,
there exist little noises in the ground
truth argument structures.
We aggregate the annotations using majority voting.

\begin{table}
\small
\centering
\begin{tabular}{c|c|c}
  \hline
\bf{Component Type}&\bf{Number} & \bf{Kappa}\\ \hline
  Major claim&143	&0.86\\ \hline
  Claim&581 &0.77	\\ \hline
  Premise&206 &0.65\\ \hline
  PSIC &121 &0.94	\\ \hline
  Background&80 &0.89\\ \hline
  Recommendation&70 &1.00\\ \hline
  Non-argumentative&145  &0.78\\
  \hline
\end{tabular}
 \caption{\label{table:corpus}The number and Fleiss' kappa for each
 argument component type we annotate.
 }
\end{table}

\section{Features}
\label{sec:features}

In line with \cite{yang2015semantic}, we consider the helpfulness
as an intrinsic characteristic of product reviews, and
thus only consider the following four intrinsic features as
our baseline features.

Structural features ({\bfseries STR})
\cite{kim2006automatically,xiong2014empirical}:
we use the following structural features: total number of tokens, total number of sentences, average length of sentences,
number of exclamation marks,
and the percentage of question sentences.

Unigram features ({\bfseries UGR})
\cite{kim2006automatically,xiong2014empirical}:
we remove all stopwords and non-frequent words (\emph{tf} $<3$)
to build the unigram vocabulary.
Each review is represented by the vocabulary with
\emph{tf-idf} weighting for each appeared term.

Emotional features ({\bfseries GALC}) \cite{martin2014prediction}:
the Geneva Affect Label Coder (GALC) dictionary proposed by \cite{scherer2005emotions} defines 36 emotion states distinguished by words. We build a real feature vector with the number of occurrences of each emotional word plus one additional dimension for the number of non-emotional words.

Semantic features ({\bfseries INQUIRER}) \cite{yang2015semantic}: the General Inquirer (INQUIRER) dictionary proposed by \cite{stone1962general} maps each word to some semantic tags, e.g. word \emph{absurd} is mapped to tags NEG and VICE; similar to the GALC features, the semantic features include
the number of occurrences of each semantic tag.

\subsection{Argument-based Features}
\label{subsec:features:argument}
The argument-based features can have different
granularity: for example, the number of argument components
can be used as features, and the number of tokens (words) in
the argument components can also be used as features.
We consider four granularity of argument features,
detailed as follows.

\textbf{Component-level argument features}.
A natural feature that we believe to be useful
is the ratio of different argument component
numbers. For example, we may be interested in
the ratio between the number of premises and
that of claims;
a high ratio suggests that there are more
premises supporting each claim, indicating that
the review gives many evidences.
To generalise this component ratio feature,
we propose \emph{component-combination ratio} features:
we compute the ratios between any
two argument components combinations.
For example, we may be interested in the ratio between
the number of MajorClaim+Claim+Premise and
that of Background+Non-argumentative.
As there are 7 types of labels, the number of possible
combinations is $2^7-1=127$, and thus the possible
number of combination ratio pairs is $127 \times 126 =
16002$. In other words, the component-level feature
is a 16002-dimensional real vector.

\textbf{Token-level argument features}. In a finer-granularity, we consider the number of
tokens in argument components to build features:
for example, suppose a review has only two claims,
one has 10 words and the other has 5 words;
we may want to know the average number of words
contained in each claim, the total number of
words in claims, etc. In total, for each
argument component type, we consider 5 types of
token-level statistics: the total number of words in the
given component type, the length
(in terms of word) of the shortest/longest component
of the given type, and the mean/variance of the
number of words in each component of the given type.
Thus, there are in total $7 \times 5 = 35$
features to represent the token-level statistics.

In addition, the ratio of some token-level
statistics may also be of interests:
for example, given a review, we may want to
know the ratio between the number of
words in Claims+MajorClaims
and that in Premises. Thus, the combination
ratio can also be applied here. We consider only the
combination ratio for two statistics:
the total number of words and the average number of
words in each component-combination;
hence, there are $16002 \times 2 = 32004$ dimensions
for the combination ratio for the statistics.
In total, there are $32004+35 = 32039$ dimensions
for the token-level argument features.

\textbf{Letter-level argument features}.
In the finest-granularity, we consider the letter-level features,
which may give some information the token-level features do not contain:
for example, if a review has a big number of
letters and a small number of words, it may suggests that
many long and complex words are used in this review,
which, in turn, may suggests that the linguistic complexity
of the review is relative high and the review may
gives some very professional opinions.
Similar to the token-level features above,
we design 5 types of statistics and their combination
ratios. Thus, the dimension for the letter-level features
is the same to that of the token-level features.

\begin{table*}
\small
\centering
\begin{tabular}{c|c|c|c|c|c}
\hline
&\bf{Accuracy}&\bf{Precision}&\bf{Recall}&\bf{F1-score}&\bf{AUC} \\
\hline
AF&0.617&0.625&0.617&0.620&0.611 \\
\hline
STR&0.600&0.360&0.600&0.450&0.500 \\
STR+AF&0.604&0.614&0.604&0.607&0.599 \\
\hline
UGR&0.697&\bf{0.760}&0.697&0.646&0.627 \\
UGR+AF&\bf{0.718}&0.718&\bf{0.719}&\bf{0.717}&\bf{0.706} \\
\hline
GALC&0.621&0.605&0.621&0.579&0.560 \\
GALC+AF&0.647&0.654&0.647&0.649&0.640 \\
\hline
INQUIRER&0.533&0.512&0.533&0.517&0.493 \\
INQUIRER+AF&0.657&0.664&0.657&0.659&0.651 \\
\hline
\end{tabular}
\caption{Helpful reviews identification performances using argument-based
features and/or baseline features.
AF stands for argument-based features. \label{table:results}}
\end{table*}

\textbf{Position-level argument features}.
Another dimension to consider argument features
is the positions of argument components:
for example, if the major claims
of a review are all at the very beginning, we may
think that readers can more easily grasp the main
idea of the review and, thus, the review is more
likely to be helpful.
For each component, we use a real number to represent
its position: for example, if a review has
10 sub-sentences (i.e. clauses) in total
and the first sub-sentence the component overlaps
is the second sub-sentence,
then the position for this component is $2/10 = 0.2$.
For each type of argument component, we may be interested
in some statistics for its positions:
for example, if a review has several premises,
we may want to know the location of the earliest/latest
appearance of premises, the average position of
all premises and its variance, etc.
Similar to the token- and letter-level features, we design
the same number of features for position-level features.

\section{Experiments}
Following \cite{o2010classification,martin2014prediction},
we model the helpfulness prediction task as a classification problem; thus, we
use accuracy, precision, recall, macro F1 and area under the curve (AUC) to as
evaluation metrics. Similar to \cite{o2010classification}, we consider a review
as helpful if and only if at least 75\% opinions for the review are positive,
i.e. $X/Y \geq 0.75$ (see X and Y in Sect. \ref{sec:corpus}).
For the features whose number of dimensions is more
than 10k (i.e. the UGR features and argument-based features), to reduce their
dimensions and to improve the performance, we only use the
positive-information-gain features in these feature sets.
In line with most existing works on helpfulness prediction
\cite{martin2014prediction,yang2015semantic}, we use the LibSVM
\cite{chang2011libsvm}
as our classifier.

The performances of different features are
presented in Table \ref{table:results}. Each number in the
table is the average performance in 10-fold cross-validation tests.
From the table we can see that, when being used together with
the argument-based features,
either of the four baseline features enjoys a performance boost
in terms of all metrics we consider.
To be more specific, in terms of accuracy, precision, recall, F1 and AUC,
the average improvement for the baseline features are
4.33\%, 10.30\%, 4.32\%, 11.01\% and 10.40\%, respectively. However, we observe that the precision of UGR+AF, although gives the second
highest score among all feature combinations, is lower than that
of UGR; we leave it for future work. Also, we notice that when using the argument-based features alone,
its performance (in terms of Precision, F1 and AUC)
is superior to those of STR, GALC and INQUIRER, and is only inferior to UGR.
However, a major drawback of
the UGR feature is its huge and document-dependent
dimensionality, while the
dimensionality of argument-based features is fixed, regardless
of the size of the input documents.
Moreover, the UGR features are sparse and problematic in online learning. To summarise, compared with the other state-of-the-art features,
argument-based features are effective in identifying helpful
reviews, and can represent some complementary information
that cannot be represented in other features.

\section{What Makes a Review Helpful ?}

Argument-based features can not only improve the performance
of review helpfulness identification, but also can be used to interpret what makes a review helpful. We analyse the information gain ranking of the argument-based features and find that, among all the positive-information-gain
argument features, 36\% are from the
token-level argument feature set, and 29\%
are from the letter-level argument feature set,
suggesting that these two feature sets are most effective
in identifying helpful reviews.
Among all the token-level argument features
with positive information gain, 69\% are ratios
of sum of token number between component-combinations,
and the remaining are ratios of the mean token numbers
between component-combinations.
We interpret this observation as follows: given a review, the larger number of tokens it contains, and the more likely
the review is helpful. In fact, helpful reviews are tend to occur in those long
reviews, which generally provide with more experiences and comments about
the product being reviewed. Among all the letter-level argument features,
around three-quarters are ratios of the sum of
the number of letters between component-combinations.
This observation, again, suggests that the length of reviews plays an important
role in the review helpfulness identification.

Moreover, among all the argument-based features with positive
information gain values, a quarter of features are the position-level argument feature. This is because the
position of each argument component influences the logic flow of reviews,
which, in turn, influences the readability, convincingness and
helpfulness of the reviews. This information can hardly
be represented by all the baseline features we considered,
and we believe this explains why the performances
of the baseline features are improved when being used
together with the argument-based features. However, among all the argument-based features with positive information gain values, only 10\% are the component-level argument feature. This indicates that compared to three finer-granularity argument features above, the component-level argument feature provides less useful information in review helpfulness identification.

\section{Conclusion and Future Work}
In this work, we propose a novel set of intrinsic
features of identifying helpful reviews, namely
the \emph{argument}-based features.
We manually annotate 110 hotel reviews,
and compare the performances of
argument-based features with those of
some state-of-the-art features.
Empirical results suggest that,
argument-based features include some complementary information
that the other feature sets do not include; as a result,
for each baseline feature, the performance
(in terms of various metrics)
of jointly using this feature and argument-based features
is higher than using this baseline feature alone.
In addition, by analysing the
effectiveness of different argument-based features,
we give some insights into which reviews are more
likely to be helpful,
from an argumentation perspective.

For future work, an immediate next step is to explore
the usage of automatically extracted arguments in
helpful reviews identification: in this work,
all argument-based features are based on manually
annotated arguments;
deep-learning based
argument mining \cite{li2017joint, eger2017neural}
has produced some promising results
recently, and we plan to investigate whether
the automatically extracted arguments can be used to
identify helpful reviews, and how the errors
made in the argument extraction stage will influence
the performance of helpful reviews identification.
We also plan to investigate the effectiveness
of argument-based features in other domains.

\section*{Acknowledgements}
Yang Gao is supported by
National Natural Science Foundation of China
(NSFC) grant 61602453,
and Hao Wang is supported by
NSFC grant 61672501, 61402447 and 61502466.

\bibliography{my}
\bibliographystyle{emnlp_natbib}

\end{document}